\documentclass{article}
\usepackage[preprint, nonatbib]{nips_2018}

\usepackage[utf8]{inputenc} 
\usepackage[T1]{fontenc}    
\usepackage{hyperref}       
\usepackage{url}            
\usepackage{booktabs}       
\usepackage{amsmath,amssymb} 
\usepackage{amsfonts}       
\usepackage{nicefrac}       
\usepackage{microtype}      
\usepackage{graphicx}
\usepackage{multirow}
\usepackage{tabu}
\usepackage{caption}

\newcommand\T{\rule{0pt}{2.9ex}}       
\newcommand\B{\rule[-1.2ex]{0pt}{0pt}} 
\newcommand{\ubold}[1]{\fontseries{b}\selectfont#1}

\usepackage{cite}

\graphicspath{{./images/}}

\title{Style Augmentation: Data Augmentation via Style Randomization}

\author{
  Philip T. Jackson,\; Amir Atapour-Abarghouei,\; Stephen Bonner, \\ \textbf{Toby Breckon,\; Boguslaw Obara} \\
  Department of Computer Science \\
  Durham University\\
  \texttt{\{p.t.g.jackson, amir.atapour-abarghouei, s.a.r.bonner,} \\ \texttt{toby.breckon, boguslaw.obara\}@durham.ac.uk}
}

\begin{document}

\maketitle

\begin{abstract}
  
  We introduce style augmentation, a new form of data augmentation based on random style transfer, for improving the robustness of Convolutional Neural Networks (CNN) over both classification and regression based tasks. During training, style augmentation randomizes texture, contrast and color, while preserving shape and semantic content. This is accomplished by adapting an arbitrary style transfer network to perform style randomization, by sampling target style embeddings from a multivariate normal distribution instead of computing them from a style image. In addition to standard classification experiments, we investigate the effect of style augmentation (and data augmentation generally) on domain transfer tasks. We find that data augmentation significantly improves robustness to domain shift, and can be used as a simple, domain agnostic alternative to domain adaptation. Comparing style augmentation against a mix of seven traditional augmentation techniques, we find that it can be readily combined with them to improve network performance. We validate the efficacy of our technique with domain transfer experiments in classification and monocular depth estimation illustrating superior performance over benchmark tasks.
  
  \end{abstract}

\section{Introduction}

Whilst deep neural networks have shown record-breaking performance on complex machine learning tasks over the past few years, exceeding human performance levels in certain cases, most deep models heavily rely on large quantities of annotated data for individual tasks, which is often expensive to obtain. A common solution is to augment smaller datasets by creating new training samples from existing ones via label-preserving transformations \cite{yaeger_effective_1997}.

Data augmentation imparts prior knowledge to a model by explicitly teaching invariance to possible transforms that preserve semantic content. This is done by applying said transform to the original training data, producing new samples whose labels are known. For example, horizontal flipping is a popular data augmentation technique \cite{krizhevsky_imagenet_2017}, as it clearly does not change the corresponding class label. The most prevalent forms of image-based data augmentation include geometric distortions such as random cropping, zooming, rotation, flipping, linear intensity scaling and elastic deformation. Whilst these are successful at teaching rotation and scale invariance to a model, what of color, texture and complex illumination variations?

Tobin et al. \cite{tobin_domain_2017} show that it is possible for an object detection model to generalize from graphically rendered virtual environments to the real world, by randomizing color, texture, illumination and other aspects of the virtual scene. It is interesting to note that, rather than making the virtual scene as realistic as possible, they attain good generalization by using an unrealistic but diverse set of random textures. In contrast, Atapour \& Breckon \cite{atapour-abarghouei_real-time_2018} train on highly photorealistic synthetic images, but find that the model generalizes poorly to data from the real world. They are able to rectify this by using CycleGAN \cite{zhu_unpaired_2017} and fast neural style transfer \cite{johnson_perceptual_2016} to transform real world images into the domain of the synthetic images. These results together suggest that deep neural networks can overfit to subtle differences in the distribution of low-level visual features, and that randomizing these aspects at training time may result in better generalization. However, in the typical case where the training images come not from a renderer but from a camera, this randomization must be done via image manipulation, as a form of data augmentation. It is not clear how standard data augmentation techniques could introduce these subtle, complex and ill-defined variations.

\begin{figure*}[t]
    \includegraphics[width=\linewidth]{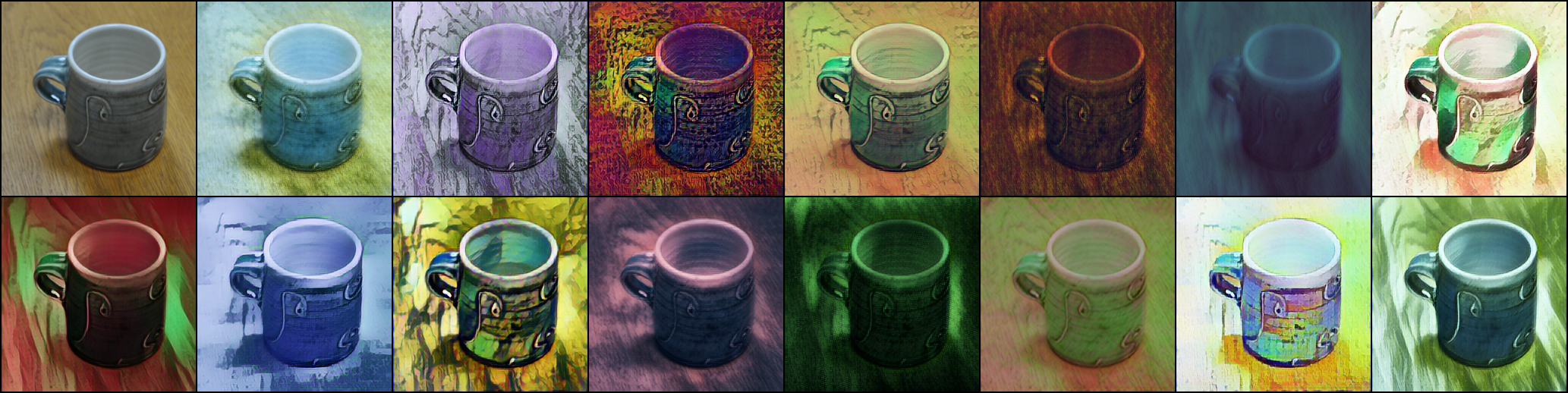}
    \caption{Style augmentation applied to an image from the Office dataset \cite{saenko_adapting_2010} (original in top left). Shape is preserved but the style, including texture, color and contrast are randomized.}
	\label{fig:mug_montage}
	\vskip -10pt
 \end{figure*}

Neural style transfer \cite{gatys_neural_2015} offers the possibility to alter the distribution of low-level visual features in an image whilst preserving semantic content. Exploiting this concept, we propose \emph{Style Augmentation}, a method to use style transfer to augment arbitrary training images, randomizing their color, texture and contrast whilst preserving geometry (see Figure \ref{fig:mug_montage}). Although the original style transfer method was a slow optimization process that was parameterized by a target style image \cite{gatys_neural_2015}, newer approaches require only a single forward pass through a style transfer network, which is parameterized by a style embedding \cite{ghiasi_exploring_2017}. This is important, because in order to be effective for data augmentation, style transfer must be both fast and randomized. Since the style transfer algorithm used in our work is parameterized by an $\mathbb{R}^{100}$ embedding vector, we are able to sample that embedding from a multivariate normal distribution, which is faster, more convenient and permits greater diversity than sampling from a finite set of style images.

In addition to standard classification benchmarks, we evaluate our approach on a range of domain adaptation tasks. To the best of our knowledge, this is the first time data augmentation has been tested for domain adaptation. Ordinarily, data augmentation is used to reduce overfitting and improve generalization to unseen images from the same domain, but we reason that domain bias \textit{is} a form of overfitting, and should therefore benefit from the same countermeasures. Data augmentation is not domain adaptation, but it can reduce the need for domain adaptation, by training a model that is more general and robust in the first place. Although this approach may not exceed the performance of domain adaptation to a specific target domain, it has the advantage of improving accuracy on \textit{all} potential target domains before they are even seen, and without requiring separate procedures for each.

In summary, this work explores the possibility of performing data augmentation via style randomization in order to train more robust models that generalize to data from unseen domains more effectively. Our primary contributions can thus be summarized as follows:

\begin{itemize} 
	\setlength\itemsep{1.0mm}
	\item \textit{Style randomization} - We propose a novel and effective method for randomizing the action of a style transfer network to transform any given image such that it contains semantically valid but random styles.
	\item \textit{Style augmentation} -	We utilize the randomized action of the style transfer pipeline to augment image datasets to greatly improve downstream model performance across a range of tasks.
	\item \textit{Omni-directional domain transfer} - We evaluate the effectiveness of using style augmentation to implicitly improve performance on domain transfer tasks, which ordinarily require adapting a model to a specific target domain post-training.
\end{itemize}%

These contributions are reinforced via detailed experimentation, supported by hyperparameter grid searches, on multiple tasks and model architectures. We open source our PyTorch implementation as a convenient data augmentation package for deep learning practitioners\footnote{URL redacted for review anonymity}.
\section{Related Work}

\subsection{Domain Bias}
\label{ssec:domain-bias}

The issue of domain bias or domain shift \cite{gretton_covariate_2008} has long plagued researchers working on the training of discriminative, predictive, and generative models. In short, the problem is that of a typical model trained on a specific distribution of data from a particular domain will not generalize well to other datasets not seen during training. For example, a depth estimation model trained on images captured from roads in Florida may fail when deployed on German roads \cite{uhrig_sparsity_2017}, even though the task is the same and even if the training dataset is large. Domain shift can also be caused by subtle differences between distributions, such as variations in camera pose, illumination, lens properties, background and the presence of distractors.

A typical solution to the problem of domain shift is transfer learning, in which a network is pre-trained on a related task with a large dataset and then fine-tuned on the new data \cite{shao_transfer_2015}. This can reduce the risk of overfitting to the source domain because convolutional features learned on larger datasets are more general \cite{yosinski_how_2014}. However, transfer learning requires reusing the same architecture as that of the pre-trained network and a careful application of layer freezing and early stopping to prevent the prior knowledge being forgotten during fine-tuning.

Another way of addressing domain shift is domain adaptation, which encompasses a variety of techniques for adapting a model post training to improve its accuracy on a specific target domain. This is often accomplished by minimizing the distance between the source and target feature distributions in some fashion \cite{long_learning_2015,ghifary_deep_2016,hoffman_simultaneous_2017,donahue_adversarial_2016,tzeng_adversarial_2017,li_revisiting_2016}. Certain strategies have been proposed to minimize Maximum Mean Discrepancy (MMD), which represents the distance between the domains \cite{long_learning_2015,sun_deep_2016}, while others have used adversarial training to find a representation that minimizes the domain discrepancy without compromising source accuracy \cite{ghifary_deep_2016,hoffman_simultaneous_2017,tzeng_adversarial_2017}. Although many adversarial domain adaptation techniques focus on discriminative models, research on generative tasks has also utilized domain transfer \cite{donahue_adversarial_2016}. Li et al. \cite{li_revisiting_2016} propose adaptive batch normalization to reduce the discrepancy between the two domains. More relevant to our work is \cite{atapour-abarghouei_real-time_2018}, which employs image style transfer as a means to perform domain adaptation based on \cite{li_demystifying_2017}.

Even though domain adaptation is often effective and can produce impressive results, its functionality is limited in that it can only help a model generalize to a specific target domain. In contrast, our approach introduces more variation into the source domain by augmenting the data (Section \ref{ssec:data-augmentation}), which can enhance the overall robustness of the model, leading to better generalization to many potential target domains, without first requiring data from them.

\subsection{Style Transfer}
\label{ssec:style-transfer}

Style transfer refers to a class of image processing algorithms that modify the visual style of an image while preserving its semantic content. In the deep learning literature, these concepts are formalized in terms of deep convolutional features in the seminal work of Gatys et al. \cite{gatys_neural_2015}. Style is represented as a set of Gram matrices \cite{schwerdtfeger_introduction_1950} that describe the correlations between low-level convolutional features, while content is represented by the raw values of high level semantic features. Style transfer extracts these representations from a pre-trained loss network (traditionally VGG \cite{simonyan_very_2014}), and uses them to quantify style and content losses with respect to target style and content images and combines them into a joint objective function. Formally, the content and style losses can be defined as:
\begin{equation}
	\mathcal{L}_c = \sum_{i \in \mathcal{C}} \frac{1}{n_i} || f_i(x) - f_i(c) ||_F^2,
	\label{eq:content-loss}\vspace{-0.1cm}
\end{equation}
\begin{equation}
	\mathcal{L}_s = \sum_{i \in \mathcal{S}} \frac{1}{n_i} || \mathcal{G}[f_i(x)] - \mathcal{G}[f_i(s)] ||_F^2,
	\label{eq:style-loss}
\end{equation}
where $c$, $s$ and $x$ are the content, style and restyled images, $f$ is the loss network, $f_i(x)$ is the activation tensor of layer $i$ after passing $x$ through $f$, $n_i$ is the number of units in layer $i$, $\mathcal{C}$ and $\mathcal{S}$ are sets containing the indices of the content and style layers, $\mathcal{G}[f_i(x)]$ denotes the Gram matrix of layer $i$ activations of $f$, and $||\cdot||_F$ denotes the Frobenius norm. The overall objective can then be expressed as:
\begin{equation}
    \min_x \mathcal{L}_c(x,c) + \lambda \mathcal{L}_s (x,s),
\end{equation}
where $\lambda$ is a scalar hyperparameter determining the relative weights of style and content loss. Originally, this objective was minimized directly by gradient descent in image space \cite{gatys_neural_2015}. Although the results are impressive, this process is very computationally inefficient, leading to the emergence of alternative approaches that use neural networks to approximate the global minimum of the objective in a single forward pass \cite{johnson_perceptual_2016,ulyanov_texture_2016,chen_fast_2016}. These are fully-convolutional networks that are trained to restyle an input image while preserving its content. Although much faster, these networks only learn to apply a single style, and must be re-trained if a different style is required, hence enabling only single-domain rather the multi-domain adaptability proposed here.%

Building on the work of \cite{ulyanov_instance_2016}, and noting that there are many overlapping characteristics between styles (e.g. brushstrokes), Dumoulin et al. \cite{dumoulin_learned_2016} train one network to apply up to 32 styles using conditional instance normalization, which sets the mean and standard deviation of each intermediate feature map to different learned values for each style. Ghiasi et al. \cite{ghiasi_exploring_2017} generalizes this to fully arbitrary style transfer, by using a fine-tuned InceptionV3 network \cite{szegedy_rethinking_2015} to predict the renormalization parameters from the style image. By training on a large dataset of style and content images, the network is able to generalize to unseen style images. Concurrently, Huang et al. \cite{huang_arbitrary_2017} match the mean and variance statistics of a convolutional encoding of the content image with those of the style image, then decode into a restyled image, while Yanai \cite{yanai_unseen_2017} concatenates a learned style embedding onto an early convolutional layer in a style transformer network similar to that of Johnson et al. \cite{johnson_perceptual_2016}.%

In this work, while we utilize the approach presented in \cite{ghiasi_exploring_2017} as part of our style randomization procedure, any style transfer method capable of dealing with unseen arbitrary styles can be used as an alternative, with the quality of the results dependent on the efficacy of the style transfer approach.

\subsection{Data Augmentation}
\label{ssec:data-augmentation}

Ever since the work of Krizhevsky et al. \cite{krizhevsky_imagenet_2017}, data augmentation has been a standard technique for improving the generalization of deep neural networks. Data augmentation artificially inflates a dataset by using label-preserving transforms to derive new examples from the originals. For example, \cite{krizhevsky_imagenet_2017} creates ten new samples from each original by cropping in five places and mirroring each crop horizontally. Data augmentation is actually a way of explicitly teaching invariance to whichever transform is used, therefore any transform that mimics intra-class variation is a suitable candidate. For example, the MNIST (handwritten digit) dataset \cite{lecun1998gradient} can be augmented using elastic distortions that mimic the variations in pen stroke caused by uncontrollable hand muscle oscillations \cite{simard_best_2003,ciresan_deep_2010}. Yaeger et al. \cite{yaeger_effective_1997} also use the same technique for balancing class frequencies, by producing augmentations for under-represented classes. Wong et al. \cite{wong_understanding_2016} compare augmentations in data space versus feature space, finding data augmentations to be superior. 

Bouthillier et al. \cite{bouthillier_dropout_2015} argues that dropout \cite{srivastava_dropout:_2014} corresponds to a type of data augmentation, and proposes a method for projecting dropout noise back into the input image to create augmented samples. Likewise, Zhong et al. \cite{zhong_random_2017} presents random erasing as a data augmentation, in which random rectangular regions of the input image are erased. This is directly analogous to dropout in the input space and is shown to improve robustness to occlusion.

The closest work to ours is that by Geirhos et al. \cite{geirhos2018imagenet}, who have recently shown that CNNs trained on ImageNet are more reliant on textures than they are on shape. By training ResNet-50 on a version of ImageNet with randomized textures (a procedure that amounts to performing style augmentation on all images), they are able to force the same network to rely on shape instead of texture. This not only agrees more closely with human behavioural experiments, but also confers unexpected bonuses to detection accuracy when the weights are used in Faster R-CNN, and robustness to many image distortions that did not occur in the training set. Our work corroborates and extends these results by showing an additional benefit in robustness to domain shift, and shows that style randomization can be used as a convenient and effective data augmentation technique.
\section{Proposed Approach}
\label{sec:approach}

For style transfer to be used as a data augmentation technique, we require a single style transfer algorithm that is both fast and capable of applying as broad a range of styles as possible. These requirements narrow our search space considerably, since most approaches are either too inefficient \cite{gatys_neural_2015} or can only apply a limited number of styles \cite{johnson_perceptual_2016, dumoulin_learned_2016}. We chose the approach of Ghiasi et al. \cite{ghiasi_exploring_2017}, for its speed, flexibility, and visually compelling results. A critical part of our data augmentation technique is providing a method for randomizing the action of the style transfer network. In this section we will introduce the style transfer pipeline we utilize and detail our novel randomization procedure.

\subsection{Style Transfer Pipeline}
\label{ssec:our-style-transfer}

\begin{figure}[t]
    \includegraphics[width=\linewidth]{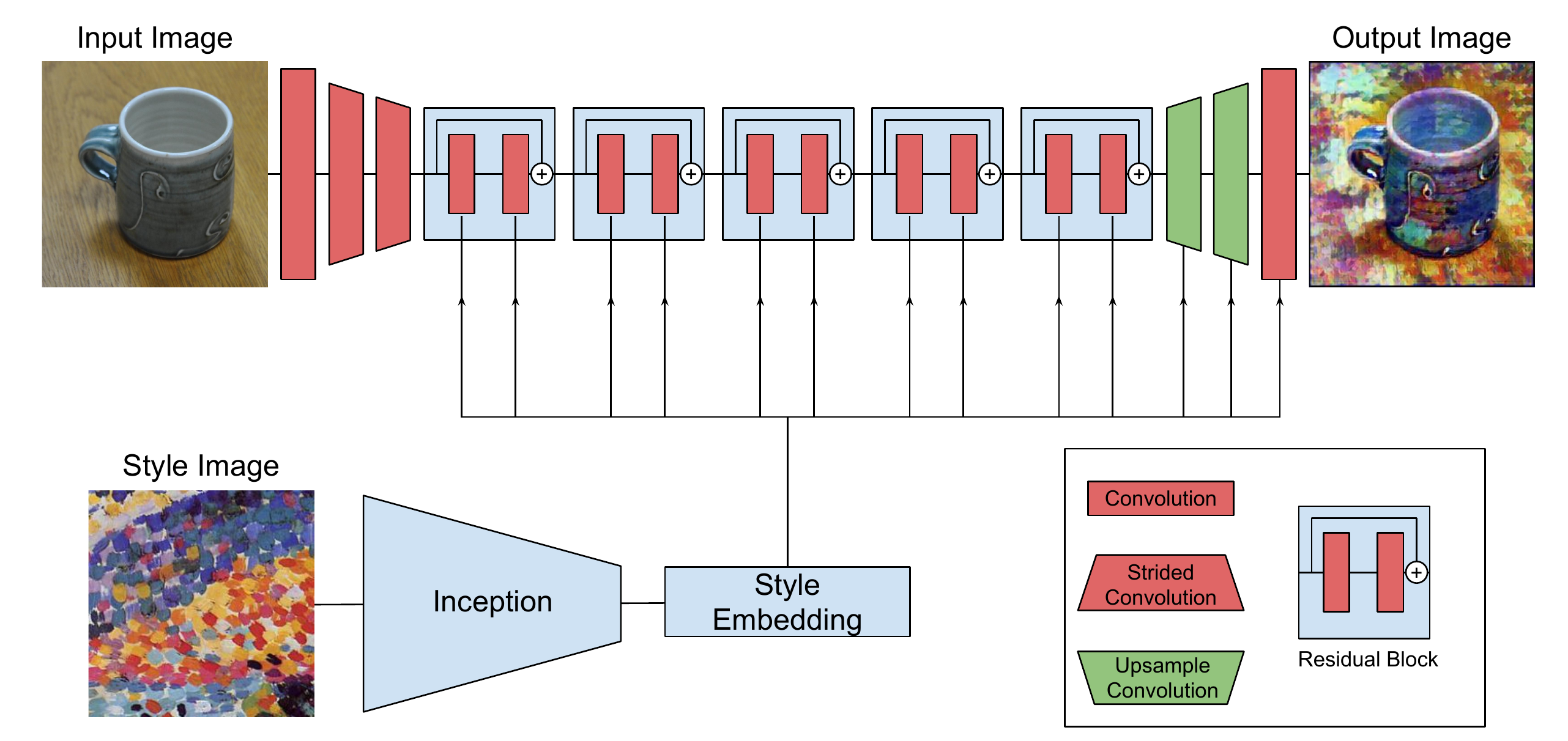}
    \caption{Diagram of the arbitrary style transfer pipeline of Ghiasi et al. \cite{ghiasi_exploring_2017}.}
	\label{fig:ghiasi_diagram}
	\vskip -10pt
\end{figure}

Our chosen style transfer network (Detailed in Figure \ref{fig:ghiasi_diagram}) employs a style predictor network to observe an arbitrary style image and output a style embedding $z \in \mathbb{R}^{100}$. For our approach we completely dispense with this style predictor network, instead we sample the style embedding directly from a multivariate normal distribution. The mean and covariance of this distribution are matched to those of the distribution of style embeddings arising from the Painter By Numbers (PBN) dataset\footnote{https://www.kaggle.com/c/painter-by-numbers}, which are used as training data for the style transfer network. Therefore, sampling from this distribution simulates choosing a random PBN image and computing its style embedding, at much lower computational cost, and without requiring the entire PBN dataset. Additionally, the size and diversity of this dataset forces the network to learn a robust mapping that generalizes well to unseen style images, much like large labelled datasets enabling classification networks to generalize well.

The style embedding $z$ influences the action of the transformer network via conditional instance normalization \cite{dumoulin_learned_2016}, in which activation channels are shifted and rescaled based on the style embedding. Concretely, if $x$ is a feature map prior to normalization, then the renormalized feature map is as follows:

\begin{equation}
	x' = \gamma (\frac{x - \mu}{\sigma}) + \beta,
	\label{eq:normalize}
\end{equation}\\
where $\mu$ and $\sigma$ are respectively the mean and the standard deviation across the feature map spatial axes, and $\beta$ and $\gamma$ are scalars obtained by passing the style embedding through a fully-connected layer. As shown in Figure~\ref{fig:ghiasi_diagram}, all convolutional layers except for the first three perform conditional instance renormalization. In this way, the transformer network output $x$ is conditioned on both the content image and the style image:

\begin{equation}
	x = T(c, P(s)).
	\label{eq:pipeline}
\end{equation}

\subsection{Randomization Procedure}
\label{ssec:style-randomization}

Randomizing the action of the style transfer pipeline is as simple as randomizing the style embedding that determines the output style. Ordinarily, this embedding is produced by the style predictor network, as a function of the given style image. Rather than feeding randomly chosen style images through the style predictor to produce random style embeddings, it is more computationally efficient to simulate this process by sampling them directly from a probability distribution. However, it is important that this probability distribution closely resembles the distribution of embeddings observed during training. Otherwise, we risk supplying an embedding unlike any that were observed during training, which may produce unpredictable behavior. We use a multivariate normal as our random embedding distribution, the mean and covariance of which are the empirical mean and covariance of the set of all embeddings of PBN images. Qualitatively, we find that this approximation is sufficient to produce diverse yet sensible stylizations (see Figure~\ref{fig:mug_montage}).

\begin{figure}[t]
	\includegraphics[width=\linewidth]{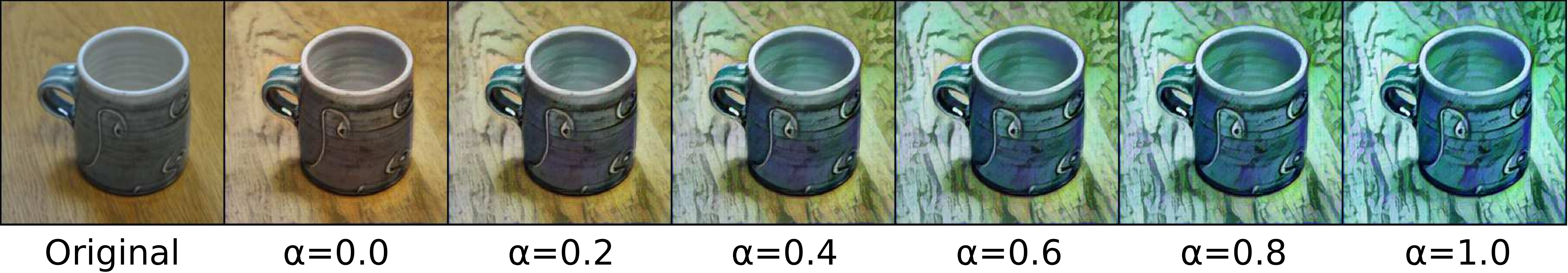}
	\caption{Output of transformer network with different values for the style interpolation parameter $\alpha$.}
	\label{fig:alpha_interpolation}
	\vskip -15pt
\end{figure}

To provide control over the strength of augmentation (see Figure~\ref{fig:alpha_interpolation}), the randomly sampled style embedding can be linearly interpolated with the style embedding of the input image, $P(c)$. Passing $P(c)$ instructs the transformer network to change the image style to the style it already has thus leaving it mostly unchanged. In general, our random embedding is therefore a function of the input content image $c$:
\begin{equation}
	z = \alpha \; \mathcal{N}(\mu,\mathbf{\Sigma}) + (1-\alpha) P(c)
	\label{eq:interpolation}
\end{equation}\\
where $P$ is the style predictor network, and $\mu$, $\mathbf{\Sigma}$ are the mean vector and covariance matrix of the style image embeddings $P(s)$:
\begin{align}
	\mu &= \mathbb{E}_s \left[ P(s) \right], \\
	\mathbf{\Sigma}_{i,j} &= \mathrm{Cov} \left[ P(s)_i, P(s)_j \right].
\end{align}

\section{Experimental Results}
\label{sec:results}

We evaluate our proposed style augmentation method on three distinct tasks: image classification, cross-domain classification and depth estimation. We present results on the STL-10 classification benchmark \cite{coates_analysis_2011} (Section~\ref{ssec:stl10}), the Office domain transfer benchmark \cite{saenko_adapting_2010} (Section~\ref{ssec:office}), and the KITTI depth estimation benchmark \cite{uhrig_sparsity_2017} (Section~\ref{ssec:depth}). We also perform a hyperparameter search to determine the best ratio of unaugmented to augmented training images and the best augmentation strength $\alpha$ (see Eqn.~\ref{eq:interpolation}). In all experiments, we use a learning rate of $10^{-4}$ and weight decay of $10^{-5}$, and we use the Adam optimizer (momentum $\beta_{1} = 0.5$, $\beta_{2} = 0.999$, initial learning rate of $0.001$).

Although we evaluate style augmentation on domain transfer tasks, our results should not be compared directly with those of domain adaptation methods. Domain adaptation uses information about a specific target domain to improve performance on that domain. In contrast, data augmentation is domain agnostic, improving generalization to all domains without requiring information about any of them. Therefore we compare our approach against other data augmentation techniques.

\begin{figure}[!t]
	\centering
	\includegraphics[width=0.9\linewidth]{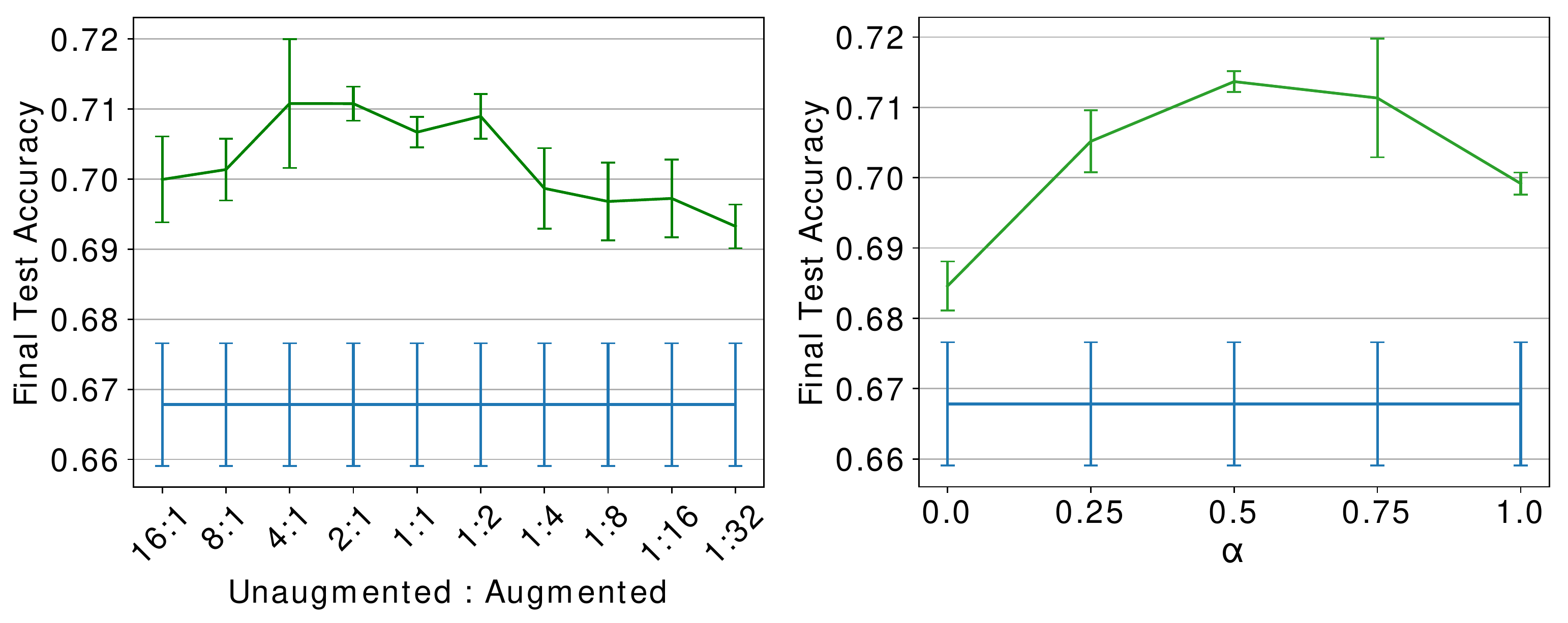}
	\caption{Hyperparameter searches on augmentation ratio and style transfer strength ($\alpha$). Curves are averaged over four experiments; error bars denote one standard deviation. Blue lines depict unaugmented baseline accuracy.}
	\label{fig:hyperparameters}
\end{figure}

\begin{table}[t!]
	\centering
	{\tabulinesep=0mm
	\begin{tabu}{@{} p{2cm} p{2cm} p{1.2cm} p{1.2cm} p{1.2cm} p{1.2cm}@{}}
	\toprule
	\multicolumn{1}{l}{\multirow{2}{*}{\textbf{Task}}}  
	& \multicolumn{1}{l}{\multirow{2}{*}{\textbf{Model}}}   
	& \multicolumn{4}{c}{\textbf{Augmentation Approach}} \\ 
	\cline{3-6}

	&& None & Trad & Style & Both \T\B \\ \hline \hline
	\multirow{4}{*}{$AW \rightarrow D$} & InceptionV3 & $0.789$  & $0.890$ & $0.882$  & $\mathbf{0.952}$ \T   \\
						   & ResNet18   &  $0.399$  & $0.704$ & $0.495$  & $\mathbf{0.873}$  \\   
						   & ResNet50   & $0.488$  & $0.778$ & $0.614$  & $\mathbf{0.922}$  \\
						   & VGG16    &  $0.558$  & $0.830$ & $0.551$  & $\mathbf{0.870}$ \B  \\ \hline
						    
	\multirow{4}{*}{$DW \rightarrow A$} & InceptionV3  & $0.183$  & $0.160$ & $0.254$  & $\mathbf{0.286}$   \T   \\
						   & ResNet18   &  $0.113$  & $0.128$ & $0.147$  & $\mathbf{0.229}$  \\
						   & ResNet50   & $0.130$  & $0.156$ & $0.170$  & $\mathbf{0.244}$  \\
						   & VGG16    &  $0.086$  & $0.149$ & $0.111$  & $\mathbf{0.243}$  \B \\ \hline

	\multirow{4}{*}{$AD \rightarrow W$} & InceptionV3  & $0.695$  & $0.733$ & $0.767$  & $\mathbf{0.884}$  \T   \\
						   & ResNet18   &  $0.414$  & $0.600$ & $0.424$  & $\mathbf{0.762}$  \\
						   & ResNet18   & $0.491$  & $0.676$ & $0.508$  & $\mathbf{0.825}$  \\
						   & VGG16    &  $0.465$  & $0.679$ & $0.426$  & $\mathbf{0.752}$  \B \\  
	\bottomrule
	\end{tabu}}
	\caption{Test accuracies on the Office dataset \cite{saenko_adapting_2010} with \textit{A}, \textit{D} and \textit{W} denoting the \textit{Amazon}, \textit{DSLR} and \textit{Webcam} domains.}
	\label{tab:office}
\end{table}

\begin{figure}[!t]
	\centering
	\includegraphics[width=0.65\linewidth]{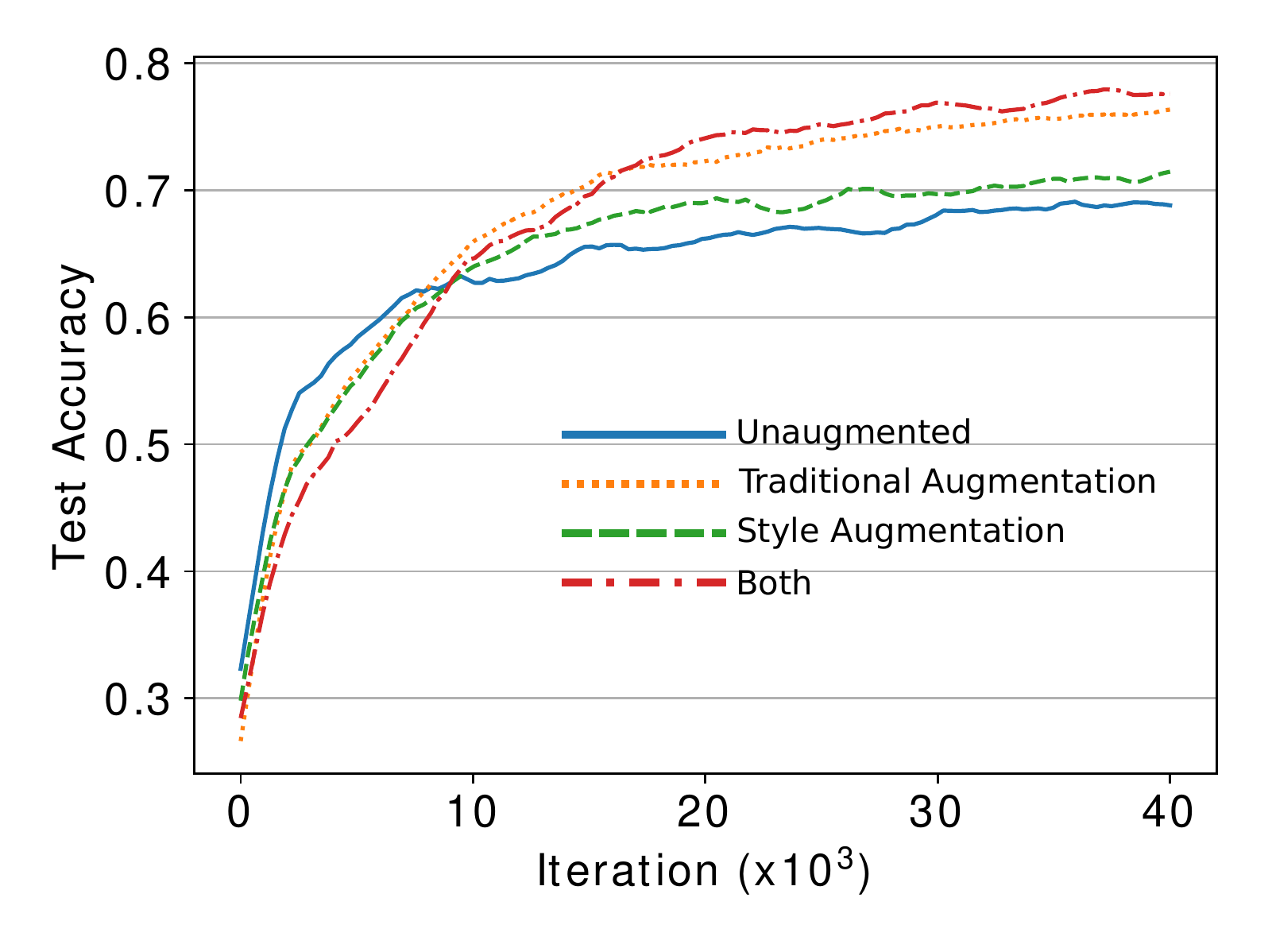}
	\caption{Comparing test accuracy curves for a standard classification task on the STL-10 dataset \cite{coates_analysis_2011}.}
	\vskip -10pt
	\label{fig:stl10-results}
\end{figure}

\subsection{Image Classification}
\label{ssec:stl10}

We evaluate our style augmentation on the STL-10 dataset \cite{coates_analysis_2011}. STL-10 consists of 10 classes with only 500 labelled training examples each, a typical case in which data augmentation would be curial since the number of labelled training images is limited. 

Prior to the final optimization, we perform a hyperparameter search to determine the optimal values for the ratio of unaugmented to augmented images and the strength of the style transfer, as determined by the interpolation hyperparameter $\alpha$. We train the InceptionV3 \cite{szegedy_rethinking_2015} architecture to classify STL-10 images, performing $40,000$ iterations, augmenting the data with style augmentation, and we repeat each experiment four times with different random seeds.

First we test augmentation ratios, interpolating in factors of two from \mbox{$16:1$} (unaugmented : augmented) to \mbox{$1:32$}. Since we do not know the optimal value of $\alpha$, we sample it uniformly at random from the interval \mbox{$[0,1]$} in these experiments. Figure~\ref{fig:hyperparameters} (left) demonstrates the results of this search. We plot the final test accuracy after $40,000$ iterations. A ratio of \mbox{$2:1$} (corresponding to an augmentation probability of $0.5$) appears to be optimal. Fixing the augmentation ratio at \mbox{$2:1$}, we repeat the experiment for $\alpha$ and find an optimal value of $0.5$ (Figure~\ref{fig:hyperparameters}, right). Style augmentation takes $2.0$ms on average per image on a GeForce 1080Ti, which corresponds to a $6\%$ training time increase on this task when the optimal augmentation ratio of \mbox{$2:1$} is used. If time is critical, the augmentation ratio can be set as low as \mbox{$16:1$} and still provide a significant accuracy boost, as Figure~\ref{fig:hyperparameters} shows. 

With suitable hyperparameters determined, we next compare style augmentation against a comprehensive mix of seven traditional augmentation techniques: horizontal flipping, small rotations, zooming (which doubles as random cropping), random erasing \cite{zhong_random_2017}, shearing, conversion to grayscale and random perturbations of hue, saturation, brightness and contrast. As in the hyperparameter search, we train InceptionV3 \cite{szegedy_rethinking_2015} to $40,000$ iterations on the $5,000$ labeled images in \mbox{STL-10}. As seen in Figure~\ref{fig:stl10-results}, while style augmentation alone leads to faster convergence and better final accuracy versus the unaugmented baseline, in combination with the seven traditional augmentations, it yields an improvement of $8.5\%$.

Moreover, without using any of the unlabeled data in STL-10 for unsupervised training, we achieve a final test accuracy of $80.8\%$ after $100,000$ iterations of training. This surpasses the reported state of the art \cite{Zhao_Stacked_2016,Thoma_2017}, using only supervised training with strong data augmentation.

\subsection{Cross-Domain Classification}
\label{ssec:office}

To test the effect of our approach on generalization to unseen domains, we apply style augmentation to the Office cross-domain classification dataset \cite{saenko_adapting_2010}. The Office dataset consists of 31 classes and is split into three domains: \textit{Amazon}, \textit{DSLR} and \textit{Webcam}. The classes are typical objects found in office settings, such as staplers, mugs and desk chairs. The \textit{Amazon} domain consists of 2817 images scraped from Amazon product listings, while \textit{DSLR} and \textit{Webcam} contain 498 and 795 images, captured in an office environment with a DSLR camera and webcam, respectively. 

We test the effect of style augmentation by training standard classification models on the union of two domains, and testing on the other. We also compare the effects of style augmentation on four different convolutional architectures: InceptionV3 \cite{szegedy_rethinking_2015}, ResNet18 \cite{he2016deep}, ResNet50 \cite{he2016deep} and VGG16 \cite{simonyan_very_2014}. For each combination of architecture and domain split, we compare test accuracy with no augmentation (None), traditional augmentation (Trad), style augmentation (Style) and the combination of style augmentation and traditional augmentation (Both). Traditional augmentation refers to the same mix of techniques as in Section~\ref{ssec:stl10}.

\begin{figure}[!t]
	\includegraphics[width=\linewidth]{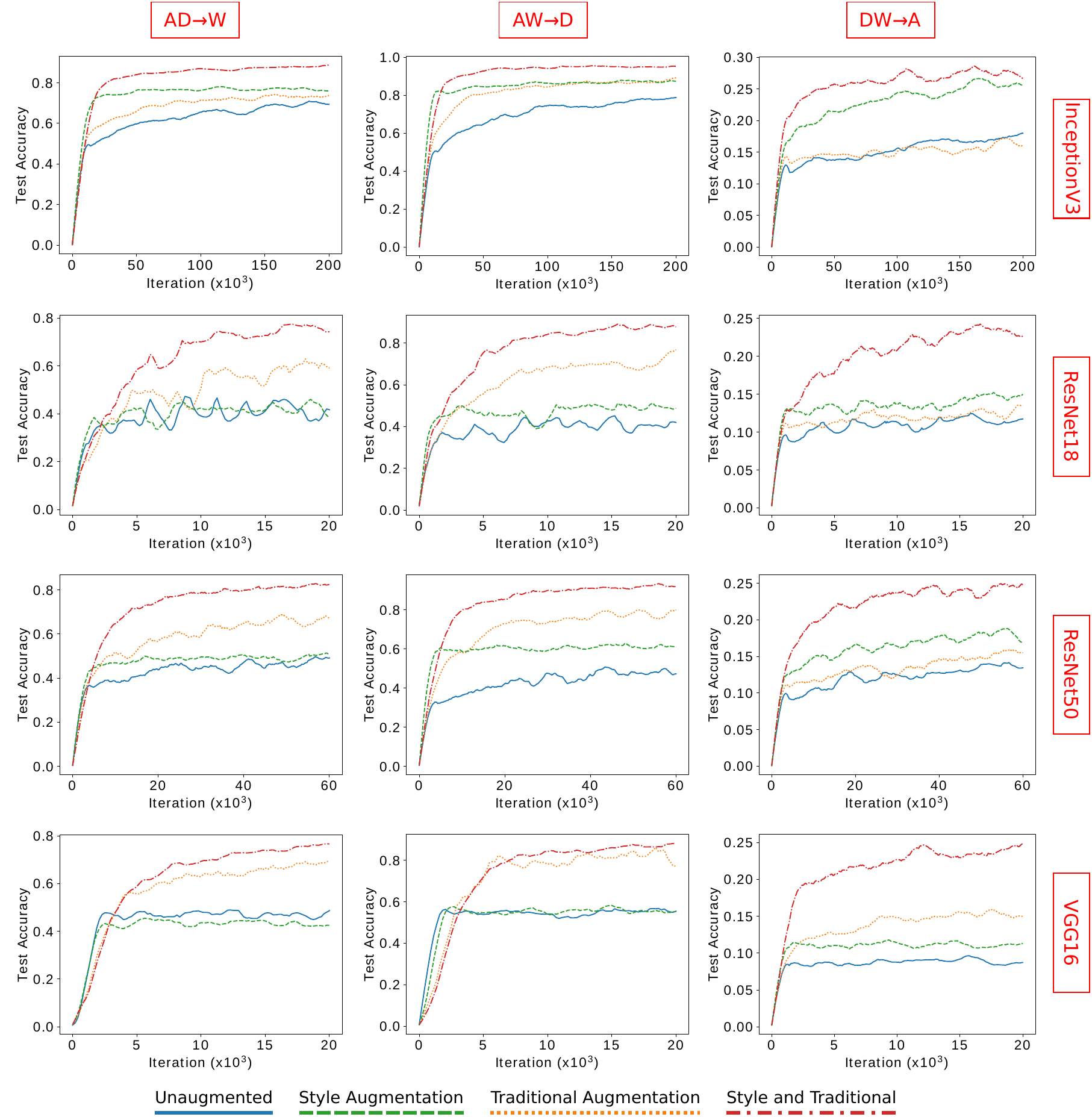}
	\caption{Results of the experiments using the Office dataset. Note the consistent superiority of traditional augmentation techniques combined with style augmentation (red curve).}
	\label{fig:office-results}
\end{figure}

Figure~\ref{fig:office-results} shows test accuracy curves for these experiments, and Table~\ref{tab:office} contains final test accuracies. In certain cases, style augmentation alone (green curve) outperforms all seven techniques combined (orange curve), particularly when the InceptionV3 architecture \cite{szegedy_rethinking_2015} is used. This points to the strength of our style augmentation technique and the invariances it can introduce into the model to prevent overfitting. 

\begin{figure*}[!h]
	\includegraphics[width=\linewidth]{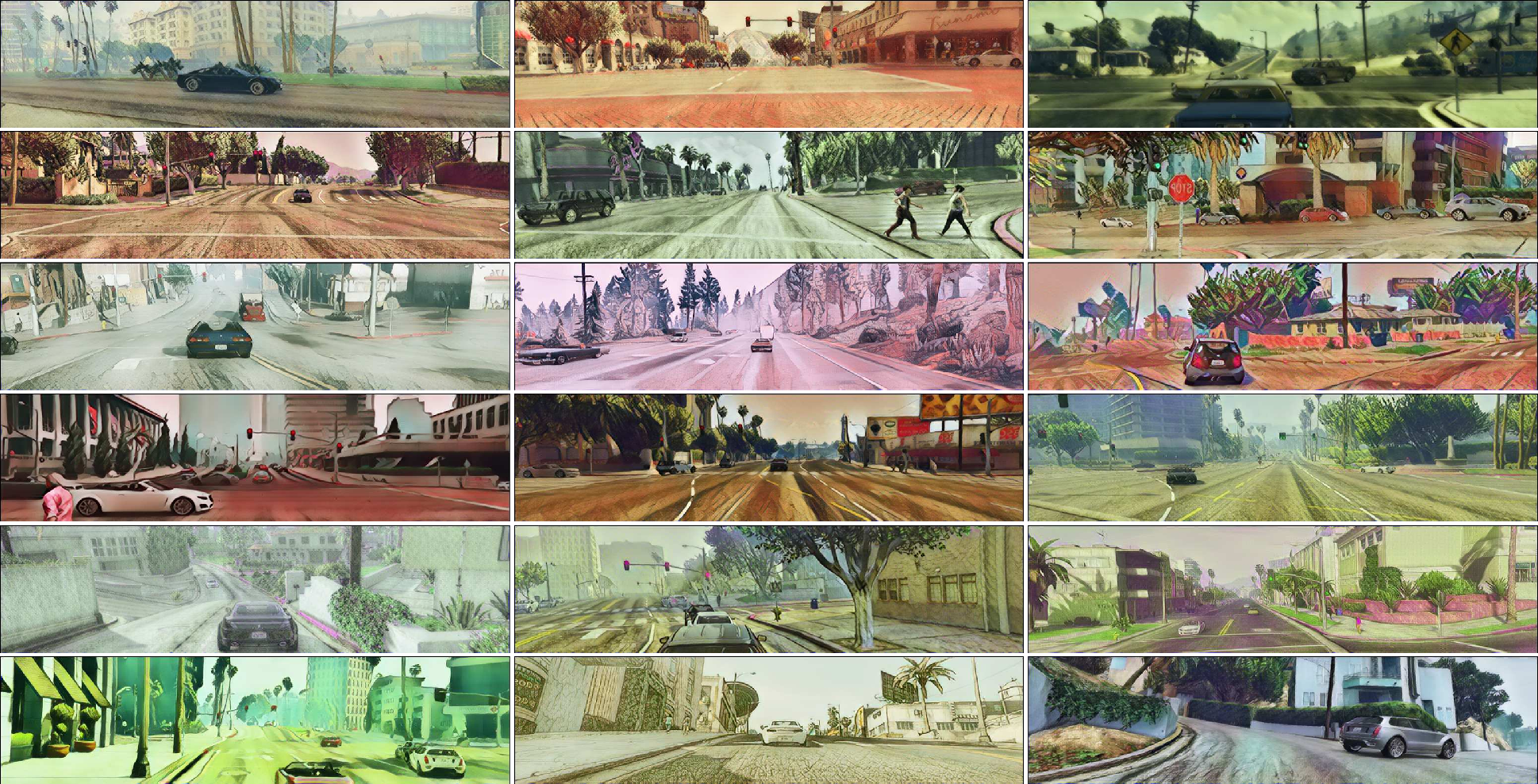}
	\caption{Examples of input monocular synthetic images post style augmentation.}
	\label{fig:style-examples}
\end{figure*}
					
An extreme domain shift is introduced into the model when the union of the \textit{DSLR} and \textit{Webcam} is used for training and the network in tested on the \textit{Amazon} domain. This is due to the large discrepancy between the \textit{Amazon} images and the other two domains and makes the classification task extremely difficult. However, as seen in Figure \ref{fig:office-results}, our style augmentation technique is capable of consistently improving the test accuracy even though the unaugmented model is barely outperforming random guess work. In all experiments, the combination of our style augmentation and traditional techniques achieves the highest final accuracy and fastest convergence (see Figure \ref{fig:office-results}).

To confirm that the benefits of style augmentation could not be realized more easily with simple colour space distortions, we ablate against color jitter augmentation, i.e. random perturbations in hue, contrast, saturation and brightness (see Table~\ref{tbl:colorjitter}). The experiment shows that style augmentation confers accuracy gains at least $4\%$ higher than those resulting from color jitter.

\begin{table}[!h]
    \centering
    \small
    \begin{tabular}{rccc}
    \toprule
     & \textbf{AD $\rightarrow$ W} & \textbf{AW $\rightarrow$ D} & \textbf{DW $\rightarrow$ A} \\ \hline \hline
    Unaugmented & 0.684 & 0.721 & 0.152 \\ 
    Color Jitter & 0.726 & 0.850 & 0.185 \\
    Style Augmentation & \textbf{0.765} & \textbf{0.893} & \textbf{0.215} \\ \bottomrule
    \end{tabular}
	\caption{Comparing style augmentation against color jitter (test accuracies on Office, with InceptionV3.)}
	\label{tbl:colorjitter}
	\vskip -10pt
\end{table}

\subsection{Monocular Depth Estimation}
\label{ssec:depth}

Finally, we evaluate our approach within monocular depth estimation - the task of accurately estimating depth information from a single image. The supervised training of a monocular depth estimation model is especially challenging as it requires large quantities of ground truth depth data, which is extremely expensive and difficult to obtain. An increasingly common way to circumvent this problem is to capture synthetic images from virtual environments, which can provide perfect per-pixel depth data for free \cite{atapour-abarghouei_real-time_2018}. However, due to domain shift, a model trained on synthetic imagery may not generalize well to real-world data.%

\begin{table*}[!h]
	\centering
	\resizebox{\textwidth}{!}{
		{\tabulinesep=0mm
			\begin{tabu}{@{\extracolsep{5pt}}c c c c c c c c@{}}
				\toprule
				\multicolumn{1}{c}{\multirow{2}{*}{\ubold{Augmentation}}} & 
				\multicolumn{4}{c}{\ubold{Error Metrics (lower, better)}} & 
				\multicolumn{3}{c}{\ubold{Accuracy Metrics (higher, better)}} \T\B \\
				\cline{2-5} \cline{6-8}
				
				& Abs. Rel. & Sq. Rel. & RMSE & RMSE log & $\sigma < 1.25$ & $\sigma < 1.25^{2}$ & $\sigma < 1.25^{3}$ \T\B \\
				
				\hline\hline
				
				None & 0.280 & 0.051 & 0.135 & 0.606 & 0.656 & 0.862 & 0.926 \T \\
				
				Trad & 0.266 & 0.045 & 0.128 & 0.527 & 0.671 & 0.872 & 0.936 \\
				
				Style & 0.256 & \ubold{0.040} & \ubold{0.123} & 0.491 & 0.696 & 0.886 & 0.942 \B \\
				
				\hline
				
				Both & \ubold{0.255} & 0.041 & \ubold{0.123} & \ubold{0.490} & \ubold{0.698} & \ubold{0.890} & \ubold{0.945} \T \B \\
				
				\bottomrule
		\end{tabu}} }
	\captionsetup[table]{skip=7pt}
	\captionof{table}{Comparing the results of a monocular depth estimation model \cite{atapour-abarghouei_real-time_2018} trained on synthetic data when tested on real-world images from \cite{uhrig_sparsity_2017}.}
	\label{table:table_depth}
\end{table*}

Using our style augmentation approach, we train a supervised monocular depth estimation network on 65,000 synthetic images captured from the virtual environment of a gaming application \cite{miralles_open-source_2017}. The depth estimation network is a modified U-net with skip connections between every pair of corresponding layers in the encoder and decoder \cite{atapour-abarghouei_real-time_2018} and is trained using a global $\ell_1$ loss along with an adversarial loss to guarantee mode selection \cite{isola_image--image_2016}. By using style augmentation, we hypothesise that the model will learn invariance towards low-level visual features such as texture and illumination, instead of overfitting to them. The model will therefore generalize better to real-world images, where these attributes may differ. Examples of synthetic images with randomized styles are displayed in Figure \ref{fig:style-examples}.

\begin{figure}[h]
	\includegraphics[width=\linewidth]{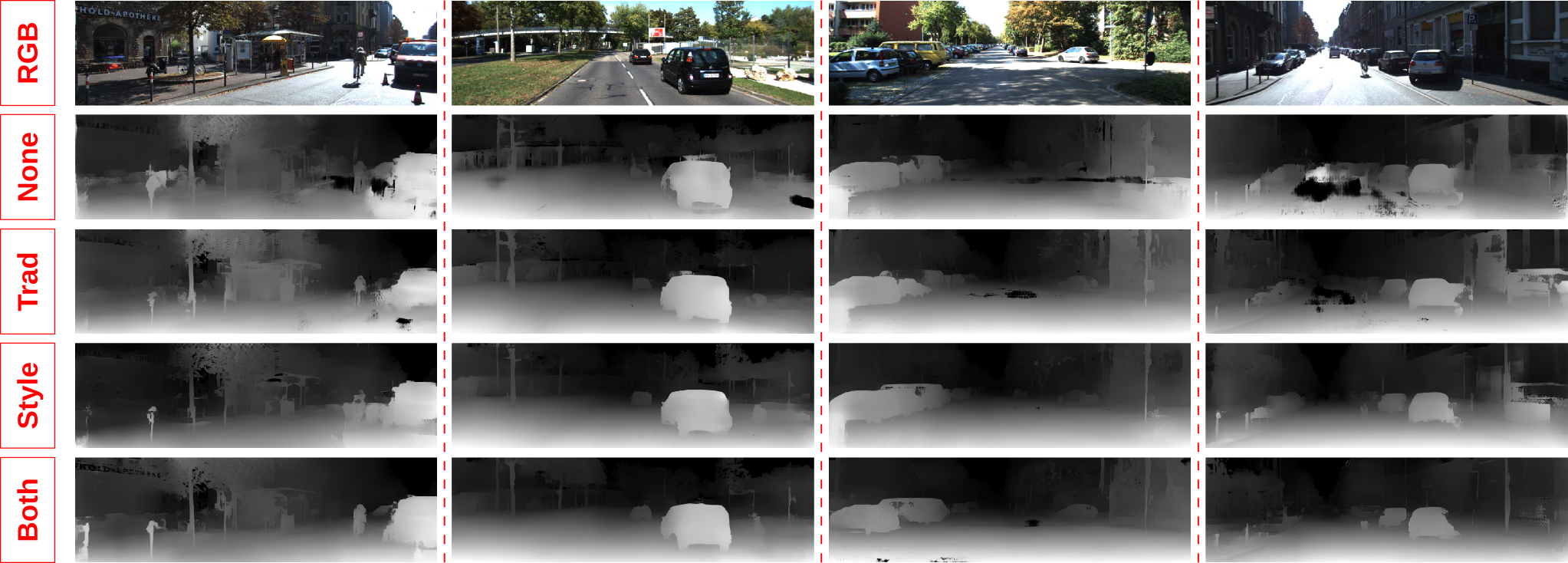}
	\caption{Results of unaugmented model (None), style (Style) traditional (None), and complete augmentation (Both) applied to depth estimation on KITTI \cite{uhrig_sparsity_2017}.}
	\label{fig:depth_eval}
	\vskip -15pt
\end{figure}

Quantitative and qualitative evaluations were run using the test split in the KITTI dataset \cite{uhrig_sparsity_2017}. Similar to our classification experiments, we compare style augmentation against traditional data augmentation techniques. However, since object scale is such a vital cue for depth estimation, any transformations that rescale the image must be ruled out. This eliminates zooming, shearing and random cropping (which requires rescaling to keep the cropped regions a constant size). Random erasing makes no sense in this context since we never estimate the depth to an occluded point. Rotation seems promising, but was empirically found to worsen the results. This leaves horizontal flipping, conversion to grayscale, and perturbations of hue, saturation, contrast and brightness as our traditional augmentations for depth estimation.

As seen in the numerical results in Table \ref{table:table_depth}, models trained with style augmentation generalize better than those trained on traditionally augmented data. These results suggest that style augmentation may be a useful tool in monocular depth estimation, given that most traditional augmentations cannot be used, and the ones that can made little difference. Moreover, qualitative results seen in Figure \ref{fig:depth_eval} indicate how our augmentation approach can produce sharper output depth with fewer artefacts.
\section{Discussion}

The information imparted to the downstream network by style augmentation, in the form of additional labelled images, is ultimately derived from the pre-trained VGG network which forms the loss function of the transformer network (see Eqn.~\ref{eq:content-loss},\ref{eq:style-loss}). Our approach can therefore be interpreted as transferring knowledge from the pre-trained VGG network to the downstream network. By learning to alter style while minimizing the content loss, the transformer network learns to alter images in ways which the content layer (i.e. a high level convolutional layer in pretrained VGG) is invariant to. In this sense, style augmentation transfers image invariances directly from pretrained VGG to the downstream network.

The case for our style augmentation method is strengthened by the work of Geirhos et al. \cite{geirhos2018imagenet}, who recently showed that CNNs trained on ImageNet learn highly texture-dependent representations, at the expense of shape sensitivity. This supports our hypothesis that CNNs overfitting to texture is a significant cause of domain bias in deep vision models, and heavily suggests style augmentation as a practical tool for combating it.

As in \cite{geirhos2018imagenet}, we found that style augmentation worsens accuracy on ImageNet - this conforms to our overall hypothesis, since texture correlates strongly enough with class label that CNNs can achieve good accuracy by relying on it almost entirely, and style augmentation removes this correlation. We do however find that style augmentation moderately improves validation accuracy on STL-10, suggesting that some image classification datasets have stronger correlation between textures and labels than others.
\section{Conclusion}
\label{sec:conclusion}

We have presented style augmentation, a novel approach for image-based data augmentation driven by style transfer. Style augmentation uses a style transfer network to perturb the color and texture of an image, whilst preserving shape and semantic content, with the goal of improving the robustness of any downstream convolutional neural networks. Our experiments demonstrate that our approach yields significant improvements in test accuracy on several computer vision tasks, particularly in the presence of domain shift. This provides evidence that CNNs are heavily reliant on texture, that texture reliance is a significant factor in domain bias, and that style augmentation is viable as a practical tool for deep learning practitioners to mitigate domain bias and reduce overfitting.

\bibliographystyle{plain}
\bibliography{New_BIB}

\end{document}